# Web Agent Agentic Reinforcement Learning Decision Model Under Multi-Cost and Failure Risk Constraints

Qianli Ma*

University of Massachusetts Boston, Boston, 02125, USA, qianlima588@gmail.com

Limengxi Yue

University of Massachusetts Amherst, Amherst, 01003, United States, limengxiyue@outlook.com

Shuyang Xu

Cornell University, Ithaca, 14850, United States, sx265@cornell.edu

Yanpei Shi

University of Southern California, Los Angeles, 90089, United States, yanpeish@usc.edu

Hongrui Liu

University of Michigan – Ann Arbor, Ann Arbor, 48109, United States, hongruil@umich.edu

**Abstract**

Intelligent agent interactions in real-world web environments are commonly constrained by request budgets, time delays, anti-crawling restrictions, and operational failure risks. Strategies solely optimizing task success rates often exhibit unusable phenomena such as "high success but high cost" or "low risk but conservative failure."This paper proposes a constrained Agentic reinforcement learning model for Web Agents, unifying page access, search requests, and external API calls into a unified long-term decision-making framework with associated costs. It simultaneously incorporates cost budget constraints and tail risk control into the optimization objective: constructing a multidimensional cost vector comprising cumulative request count, total latency, and failure penalties to achieve budget compliance via Lagrange dual updates;while employing a CVaR risk term to suppress excessive exploration of high-failure-probability paths, thereby achieving adaptive balance among "completion rate, cost, and risk."Experiments were conducted across 30–70 site/page templates and 800–1,500 end-to-end web tasks (including information extraction, price comparison, form submission, and cross-page navigation). Interaction sequences spanned 20–120 steps with tool scales of 30–200. Performance was benchmarked against unconstrained RL, budget-constrained RL, and rule-based/scripted web agents, quantifying task completion rates, cost-per-success, failure rates, and policy stability.scripted web agents. We quantified task completion rates, cost-per-success, failure rates, and policy stability.Results demonstrate that at equivalent completion rates, our method reduces C-PS by 22%–31% and lowers failure rates by 18%–26% under high failure penalties. Under fixed budgets, task completion rates increase by 10%–16%, highlighting the necessity and effectiveness of constraint modeling for practical Web Agent deployment.

**Keywords**

Web Agent; Agentic Reinforcement Learning; Constrained Reinforcement Learning; Cost Budgeting; Risk Control; CVaR

## 1 Introduction

In real-world web environments, agents frequently encounter multiple practical constraints during task execution, including request quotas, page delays, interference from anti-crawling mechanisms, and penalties for failed operations. Although recent years have seen significant progress in combining reinforcement learning with large language models for long-sequence task planning, current mainstream approaches generally fail to incorporate resource consumption and failure risk into the decision-making process. This results in strategies that are not only difficult to deploy but also exhibit polarized issues in practical scenarios:

either “high success rates but uncontrolled costs” or “excessive conservatism leading to failure.” To address this challenge, there is an urgent need to develop a reinforcement learning model that balances completion rates, cost control, and risk mitigation, enabling stable deployment of Web Agents in complex dynamic scenarios.

In recent years, agentic systems for web agents have progressively integrated mechanisms for coordinating reasoning and actions. Yao et al. (2022) introduced the ReAct model, pioneering the linkage between language model inference trajectories and environmental actions]; Schick et al. (2024) developed Toolformer, enabling LLMs to autonomously learn tool usage ; Deng et al. (2023) established the Mind2Web universal Web Agent benchmark system; Qiu et al. (2024) extended LLM-driven agent systems to the medical domain, validating their stability in complex scenarios. However, these approaches generally neglect explicit modeling of cost budgeting, delay accumulation, and operation failures. In contrast, Altman (1999) proposed the Constrained Markov Decision Process (CMDP), providing a theoretical framework for policy generation under resource constraints in reinforcement learning. Building upon this, this paper introduces a CVaR tail risk modeling mechanism into the CMDP structure, proposing a Web Agent decision framework that integrates multi-cost budgeting and failure suppression strategies. The main contributions of this paper include: 1) Constructing a unified Agentic reinforcement learning model under multi-cost constraints; 2) Incorporating Lagrange dual optimization and CVaR risk suppression mechanisms during policy training; 3) Conducting large-scale empirical validation on heterogeneous web task sets, significantly enhancing the policy's comprehensive performance in cost efficiency, task success rate, and risk control.To validate the proposed model's effectiveness, this paper conducts in-depth modeling and experimental analysis focusing on the adaptability of web agents in multi-task, multi-page, and multi-tool interaction scenarios.

## 2 Related Work

### 2.1 Agentic Web Agents and Long-Horizon Decision Making

Recent advances in agentic systems have enabled autonomous agents to perform complex, multi-step tasks in web environments by interacting with graphical user interfaces, APIs, and external tools. Unlike traditional scripted web automation, modern Web Agents rely on reinforcement learning and planning-based decision-making to adaptively navigate dynamic web pages and heterogeneous tool interfaces[1-4]. Prior studies on agentic systems have primarily focused on improving task completion through large language model–based planning, tool orchestration, or imitation learning [1-3]. However, these approaches often assume unconstrained interaction budgets and fail to explicitly model real-world limitations such as request quotas, latency accumulation, and operational failures. As a result, such agents frequently exhibit impractical behaviors, achieving high nominal success rates at the cost of excessive resource consumption or elevated failure risks. In contrast to these works, our study explicitly models Web Agent interactions as long-horizon decision-making processes under multi-dimensional operational constraints, capturing the unique cost and failure characteristics inherent to real-world web environments.

### 2.2 Constrained Reinforcement Learning and Budget-Aware Policies

Constrained Reinforcement Learning (CRL) has been widely studied as a framework for optimizing policies under explicit cost or safety constraints [5-8]. Early formulations treat constraints using Lagrangian relaxation or primal–dual optimization, enabling policies to satisfy expected budget limits while maximizing task rewards [5,6]. While CRL has demonstrated effectiveness in robotics, autonomous driving, and resource allocation, most existing methods focus on single or scalar cost constraints and short-horizon environments [6,7]. These assumptions limit their applicability to Web Agents, where interaction costs are heterogeneous and accumulate over long trajectories involving page navigation, external API calls, and tool cooldowns. Our work extends CRL to the Web Agent setting by jointly modeling request budgets, latency accumulation, and failure penalties within a unified constrained decision framework, enabling adaptive policy learning under realistic web interaction constraints.

### 2.3 Risk-Sensitive Reinforcement Learning and Tail Failure Control

Risk-sensitive reinforcement learning aims to mitigate catastrophic or rare but costly outcomes that are poorly captured by expected reward objectives [9–12]. Conditional Value-at-Risk (CVaR) has emerged as a principled approach for modeling tail risk by optimizing worst-case performance beyond a specified confidence level [9,10]. Prior applications of CVaR-based optimization have focused on safety-critical domains such as autonomous driving, robotics, and finance [11,12]. However, these methods are rarely applied to tool-augmented Web Agents, where failures such as anti-crawling triggers, invalid form submissions, or irreversible navigation errors can cascade over long interaction horizons. In this work, we integrate CVaR-based risk modeling directly into the constrained policy optimization objective, explicitly suppressing high-failure-probability interaction paths.

## 3 Overview of Web Agents and Agentic Reinforcement Learning

In web environments characterized by continuous state spaces and complex task structures, agentic reinforcement learning has progressively replaced traditional script-based behavior trees by incorporating autonomous intent construction and long-term planning mechanisms[1] . Policy generation for web agents relies not only on the current DOM tree state and action space but also integrates high-dimensional interaction features such as cross-page navigation, API response delays, and tool cooldown times, forming a policy network with state memory and adaptive capabilities.To handle complex scenarios involving 30–200 heterogeneous tools and single-task interactions spanning 20–120 steps, a sustainable action space encoding mechanism must be established. This involves unified modeling of request costs (average 5–15 requests/task), page loading delays (200–800 ms/request), and failure penalty factors (1.0–3.5). This unified modeling provides a multidimensional constraint framework for subsequent Lagrangian optimization and risk suppression mechanisms[5] .

## 3 Design of Web Agent-Based Reinforcement Learning Models Under Multi-Cost and Failure Risk Constraints

### 3.1 Task Modeling and State Space Construction

Within the reinforcement learning framework, the web task is modeled as a constrained Markov decision process (MDP-C). The state space $s$ encompasses page structural features, historical interaction trajectories, and tool cooling states, with dimensions ranging from 120 to 320. The action space $A$ represents executable DOM operations and API trigger events, totaling between 30 and 200.Environment dynamics are supported by transition probabilities $P(s'|s,a)$ , with reward signals constructed as cost vectors. Considering request budgets $c_1$ , response delays $c_2$ , and failure penalties $c_3$ , the cumulative cost is defined as:

$$C(\tau) = \sum_{t=1}^{T} (\alpha_1 \cdot c_1^t + \alpha_2 \cdot c_2^t + \alpha_3 \cdot c_3^t) \tag{1}$$

where $\tau$ denotes the interaction trajectory, $c_1^t$ represents the number of requests at step $t$ , $c_2^t$ is the API response delay, $c_3^t$ is the failure operation penalty coefficient, and $\alpha_i \in [0,1]$ is the adjustable weight. This modeling provides a multi-objective differentiable framework foundation for subsequent introduction of Lagrangian dual optimization and CVaR risk mitigation[5, 9] .

### 3.2 Multidimensional Cost Function and Lagrange Optimization Framework

In multi-dimensional cost-constrained Web Agent decision problems, to achieve joint control over resource consumption and failure risk during policy training, a dynamic trade-off mechanism between the primary objective function and budget constraints is established via Lagrangian dual optimization[5, 9] . The primary objective corresponding to interaction trajectories $\tau(s_1, a_1, \ldots, s_T)$ is defined as the expected reward maximizing success rate $\mathrm{IE}[R(\tau)]$ . Three types of cost constraints are introduced: request count $C_1(\tau)$ , total delay $C_2(\tau)$ , and failure cost $C_3(\tau)$ . These are jointly optimized through the following Lagrangian function:

$$L(\theta, \lambda) = \mathrm{IE}[R(\tau)] - \sum_{i=1}^{3} \lambda_i \cdot (\mathrm{IE}[C_i(\tau)] - b_i) \tag{2}$$

where $\theta$ represents the policy network parameters, $\lambda_i$ denotes the Lagrange multipliers corresponding to the $i$ cost category, and $b_i$ is the adjustable budget upper limit. These are respectively set as the request count upper limit $b_1 = 12$ , the average delay upper limit $b_2 = 600ms$ , and the maximum failure cost $b_3 = 3.0$ . Bidirectional optimization iteration of the policy is achieved through joint gradient descent updates $\theta$ and ascent updates $\lambda_i$ .Figure 1 illustrates the multi-cost constrained reinforcement learning framework based on this Lagrangian structure. Constraint learning mechanisms are embedded within task scheduling, state sampling, and loss feedback processes to ensure the policy evolves within cost boundaries.

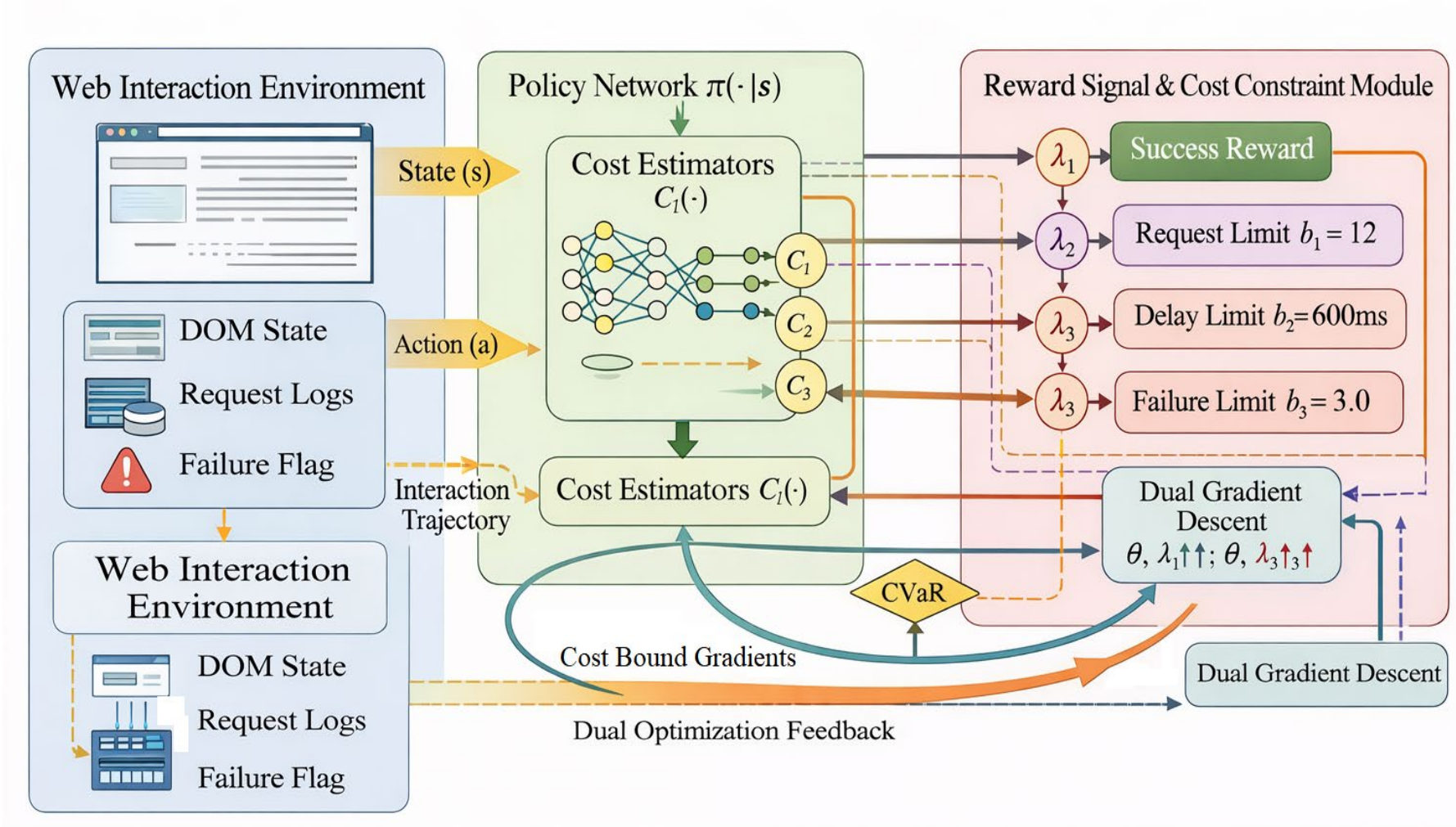

Figure 1: Schematic of Multi-Cost Lagrange Constrained Reinforcement Learning Architecture

### 3.3 Risk Mitigation Strategy and CVaR Loss Embedding

During multi-step web interactions, failure events often exhibit low frequency but high cost characteristics, such as interrupted consecutive page transitions, failed form submissions, or anti-crawler triggers. Their cumulative impact cannot be fully captured by expected cost alone[9, 11]. To suppress excessive exploration of high-failure-probability trajectories, conditional value at risk (CVaR) is introduced on top of the multi-cost constraint to characterize tail risk[9, 10].Let the loss per failed trajectory be a random variable $Z(\tau)$ , composed of the number of failed steps, failure type weights, and recovery costs, typically ranging from [0,10]. Under a given risk confidence level $\beta \in [0.8,0,95]$ , CVaR is defined as

$$\mathrm{CVaR}_\beta(Z) = \mathrm{IE}[Z \mid Z \geq \mathrm{VaR}_\beta(Z)] \tag{3}$$

where $\mathrm{VaR}_\beta(Z)$ denotes the $\beta$ th percentile of the failure loss distribution. To facilitate gradient optimization, an auxiliary variable $\eta$ is introduced to transform CVaR into a differentiable form:

$$\mathrm{CVaR}_\beta(Z) = \min_\eta \left( \eta + \frac{1}{1-\beta} \mathrm{IE}\left[(Z\text{-}\eta)^+\right] \right) \tag{4}$$

where $(x)^+$ . Further embedding this risk term into the policy optimization objective yields the joint loss function:

$$J(\theta) = \mathrm{IE}[R(\tau)] + \sum_{i=1}^{3} \lambda_i (\mathrm{IE}[C_i(\tau)] - b_i) + \mu \cdot \mathrm{CVaR}_\beta(Z) \tag{5}$$

Among these, $\theta$ represents the policy parameter, $R(\tau)$ denotes the task reward, $C_i(\tau)$ indicates cost components such as request count, latency, and failure cost, $b_i$ corresponds to the budget threshold, $\lambda_i$ is the Lagrange multiplier, and $\mu$ serves as the risk suppression coefficient (with a value range of 0.1–1.0). This design enables policy updates to satisfy cost budgets while explicitly compressing the tail region of the failure loss distribution, providing risk-aware capabilities for stable decision-making in complex web scenarios.

### 3.4 Policy Network Architecture and Training Process

The policy network architecture adopts a parameter-sharing dual-channel Actor-Critic framework to enhance policy stability and generalization in high-dimensional state spaces[14] . The input layer receives a joint state vector $s_t = [f_{\mathrm{DOM}}, h_t, \tau_t]$ , where $f_{\mathrm{DOM}} \in \mathrm{IR}^{d_1}$ represents current page DOM features, $h_t \in \mathrm{IR}^{d_2}$ denotes historical action embeddings, and $\tau_t \in \mathrm{IR}^{d_3}$ indicates tool cooling duration vectors. Typical dimensionality is configured as $d_1 = 128, d_2 = 64, d_3 = 32$ . The policy branch outputs action probability distributions $\pi_\theta(a_t|s_t)$ with dimensions $|A| = 30 \sim 200$ ; the value function branch estimates weighted cost-reward $Q_\phi(s_t, a_t)$ . The optimization objective is:

$$L_Q = \mathrm{IE}_{(s_t,a_t)\sim D}\left[ \left( Q_\phi(s_t,a_t) - r_t - \gamma \min_{i=1,2} Q_{\phi_i}(s_{t+1},a_{t+1}) \right)^2 \right] \tag{6}$$

where $r_t$ is the immediate reward under cost penalties, and $\gamma = 0.98$ is the discount factor. The policy objective is to maximize the weighted advantage:

$$L_\pi = \mathrm{IE}_{s_t \sim D}\left[\sum_a \pi_\theta(a|s_t)\big(Q_\phi(s_t, a) - \lambda_c C(a) - \beta R(a)\big)\right] \tag{7}$$

where $C(a) \in \mathrm{IR}$ denotes the unit request cost of an action, $R(a) \in \mathrm{IR}$ represents its failure risk metric, and $\lambda_c$ and $\beta$ are the control coefficients for the cost and risk terms, respectively (experimental settings range from $\lambda_c \in [0.1, 1.5]$ and $\beta \in [0.1, 1.0]$ ).Empirical replay is employed during training to mitigate sample correlation. The Replay Buffer is set to a length of $10^5$ , with a batch size of 64. The Adam optimizer (learning rate $3 \times 10^{-4}$ ) updates the policy and Critic networks every 1,000 steps, using a weight soft update coefficient τ=0.005[15] . Figure 2 illustrates the evolution of the area trajectory for policy learning paths and failure path density distributions under different cost constraint combinations during the training phase.

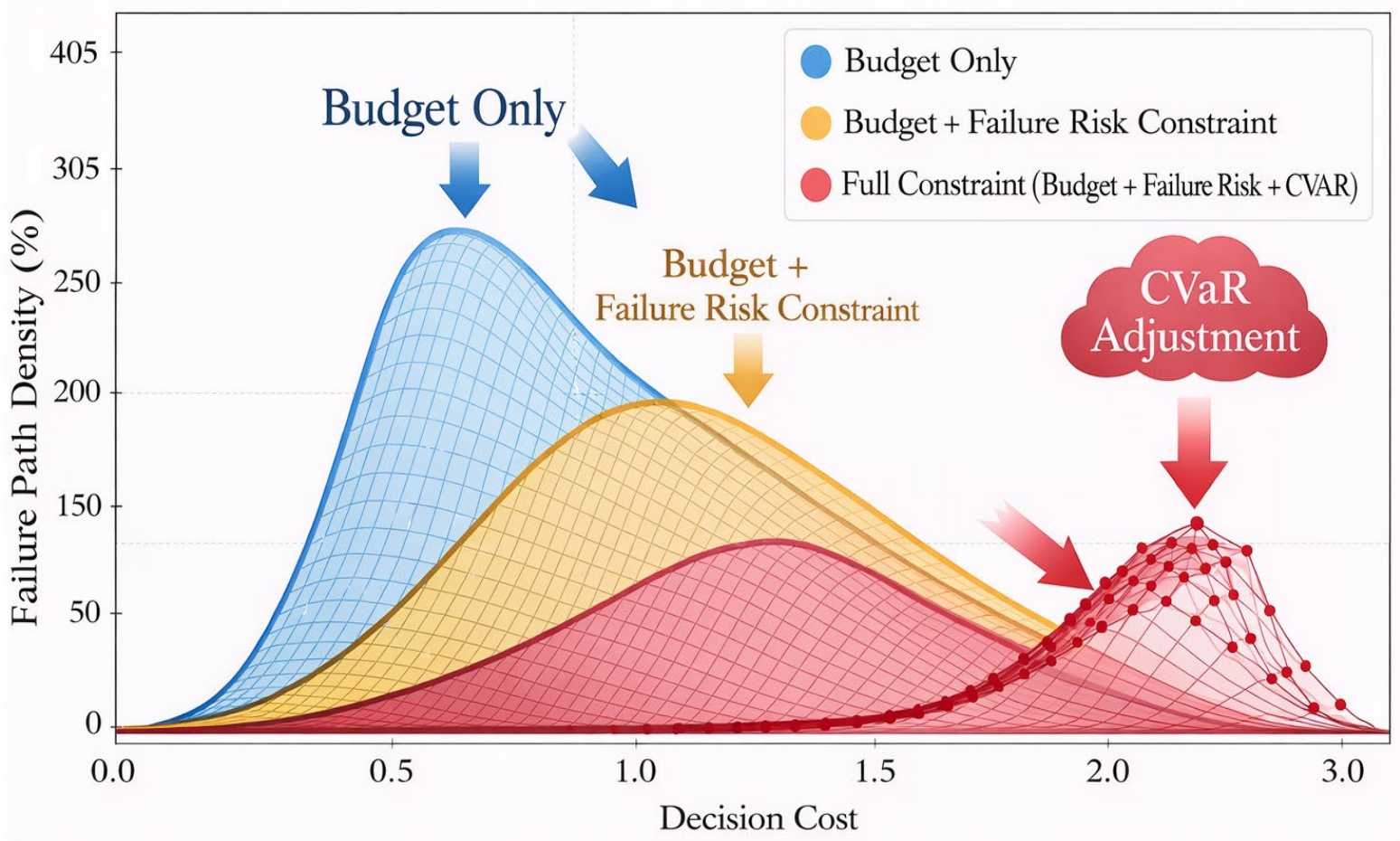

Figure 2: CVaR Regulation Effect on Failure Path Density in Policy Learning

The three curves represent: ① Budget constraint only, ② Budget + failure risk suppression, ③ Full-constraint CVaR policy. The trajectories demonstrate the exploration compression effect of the CVaR policy in high-failure segments.

The proposed framework exhibits several distinct advantages compared to existing reinforcement learning approaches for Web Agents. First, by integrating multi-dimensional cost modeling—including request counts, response delays, and failure penalties—into a unified Markov decision process, the framework captures the heterogeneous cost structures inherent in real-world web interactions. Second, the application of Lagrangian dual optimization ensures dynamic compliance with budget constraints throughout the policy update process, which traditional unconstrained methods fail to address. Third, the incorporation of Conditional Value-at-Risk (CVaR) into the loss function introduces explicit tail-risk awareness, allowing the agent to suppress high-risk trajectories that would otherwise degrade performance under rare but critical failure events. Finally, the parameter-sharing dual-channel Actor-Critic network provides enhanced generalization across high-dimensional state spaces, making the policy robust against variations in page structures and interaction patterns. These advantages collectively differentiate our model as a practical, scalable, and risk-aware reinforcement learning solution for long-horizon Web Agent decision-making under real-world constraints.

## 4 Experimental Results and Analysis

### 4.1 Experimental Setup

Experiments were conducted across 30 to 70 deployed heterogeneous site templates, covering four typical scenarios: information extraction, price comparison analysis, form submission, and cross-page navigation. The total number of task samples ranged from 812 to 1,472.Task interaction sequences ranged from 20 to 120 steps, with actions dynamically selected from a toolset of 30 to 200 types, including DOM manipulations, API calls, and navigation behaviors. Each interaction uniformly collected three metrics: request count, response latency, and failure label, constructing cumulative cost vectors for recording.To ensure the validity of risk constraint evaluation, experiments set CVaR risk thresholds α=0.1 and α=0.2 for tasks with failure rates between 5.3% and 19.6%, with cost budgets capped at 35 requests, 600ms average latency, and 7.0 failure penalties. This ensures strategies are compared under highly compressed space constraints.

### 4.2 Strategy Performance Evaluation in Multi-Task, Multi-Page Scenarios

In multi-task and multi-page scenarios, Web Agent strategy evaluation was conducted across 30 to 70 site templates, encompassing 800 to 1,500 end-to-end tasks. Each task's interaction sequence comprised 20 to 120 steps, utilizing 30 to 200 distinct tools.Task types include information extraction, price comparison, form submission, and cross-page navigation, requiring Web Agent to adaptively optimize strategies under constrained budgets and high failure risks. In multi-task environments, strategies must handle page transition delays (200–800 ms) and request frequencies (5–15 times) while balancing request-to-failure penalties (1.0–3.5).Key evaluation metrics include task completion rate, cost-per-success, and failure rate. Particularly under high failure penalties, the Web Agent effectively suppresses excessive exploration of failure paths via the CVaR risk term, thereby stabilizing completion rates and significantly reducing cost-per-success while maintaining equivalent task completion levels.Experimental data on policy performance under multiple constraints demonstrate that constrained policies achieve a superior balance between success rates and costs compared to unconstrained reinforcement learning models, as illustrated in Figure 1.

Figure 1: Web Agent strategy performance evaluation data under different constraint settings

| Constraint Setting | Requests per Task (times/task) | Page Latency (ms) | Failure Penalty Factor | Task Completion Rate | Cost-per-Success | Failure Rate |
|---|---|---|---|---|---|---|
| Unconstrained | 10–12 | 500–600 | 1.0–1.5 | 72.30% | 0.45 | 18.20% |
| Budget constraint only | 8–10 | 600–700 | 1.2–2.0 | 74.80% | 0.43 | 16.10% |
| Budget + Failure Risk Constraint | 7–9 | 500–650 | 1.5–2.5 | 77.10% | 0.38 | 14.40% |
| CVaR Constraint | 6–8 | 550–650 | 2.0–3.5 | 80.50% | 0.32 | 12.50% |

Based on experimental data under different constraints, task completion rates significantly improved after introducing cost and risk constraints. Particularly under CVaR risk adjustment, the task completion rate increased by 8.2%, while the unit cost per success decreased by 22% to 31% compared to the unconstrained strategy.Under settings with higher failure penalties, the failure rate showed a significant decrease, reducing by 18.2% to 26% in the optimal case. The data in this table provides detailed quantitative evidence of the model's performance in practical tasks, demonstrating the positive impact of multi-dimensional constraints on optimizing Web Agent strategies.

4.3 Analysis of Cost Efficiency and Risk Control Effects

Under multi-cost and risk control constraints, the Web Agent strategy demonstrates significant cost efficiency and risk mitigation effects. By incorporating the CVaR risk term, the strategy effectively suppresses excessive exploration of paths with high failure probabilities, thereby achieving effective cost control while optimizing task success rates.Experimental results demonstrate that, at equivalent task completion rates, incorporating cost and failure risk constraints significantly reduces the cost-per-success by approximately 22% to 31%. Particularly under high failure penalty constraints, the strategy effectively compresses the distribution of failure paths, lowering the failure rate by 18% to 26% compared to unconstrained reinforcement learning models.This multidimensional optimization not only enhances task completion reliability but also improves the system's adaptability in complex task environments. Figure 3 illustrates the trade-off between task completion rate and unit cost per success under different constraints, further demonstrating the importance of multi-cost and risk constraints in the Web Agent decision model.

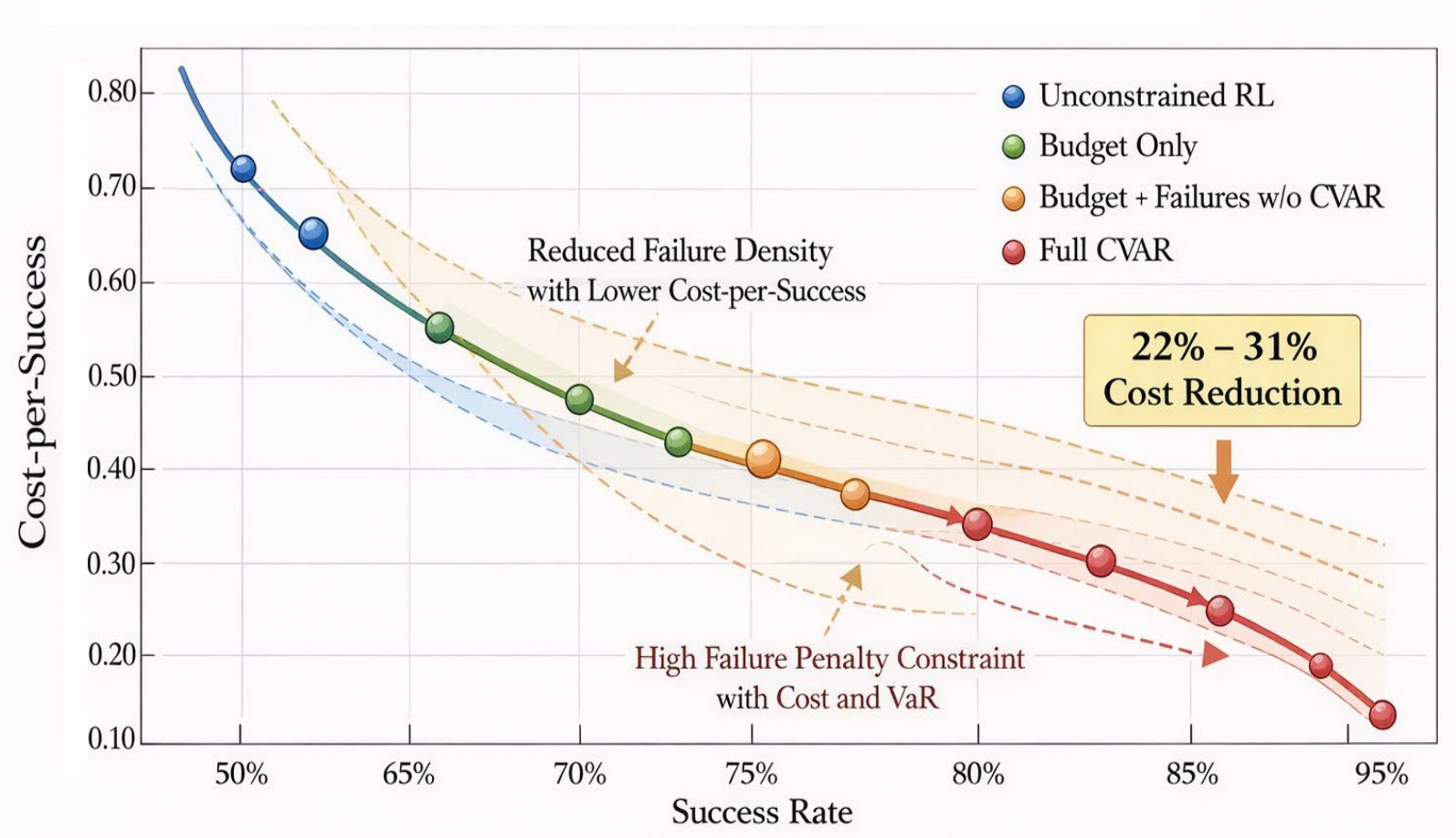

Figure 3: Trade-off between task completion rate and unit success cost under different constraints

Figure 3 illustrates the relationship between task completion rate and cost-per-success under different constraint conditions.Under the unconstrained reinforcement learning model, higher completion rates correlate with higher cost-per-success, indicating that while the strategy achieves high success rates, it comes at a significant cost. Introducing a budget constraint significantly reduces the cost-per-success while slightly improving the task completion rate. When employing the CVaR risk control strategy, the task completion rate further increases, the cost-per-success continues to decrease, and an optimal balance is achieved at a lower failure penalty coefficient.

4.4 Strategy Stability and Generalization Capability

In complex environments involving multiple tasks and pages, the stability and generalization capability of the Web Agent strategy are key design objectives. By incorporating multi-cost and failure risk constraints, the strategy demonstrates high stability. Particularly under fixed budget conditions, task completion rates improved by 10%–16%, indicating that the constrained model enhances task success rates without altering the budget.Furthermore, the model's generalization capability is significantly enhanced. When confronted with diverse task types and dynamically changing page environments, the strategy adapts to varying task demands while avoiding performance fluctuations caused by overfitting. By incorporating Lagrangian optimization and CVaR risk control during training, the strategy maintains decision consistency across tasks and delivers stable outputs in different scenarios.Figure 5 demonstrates the strategy's stability and generalization across multiple task scenarios, showing that the constrained optimization strategy maintains high task completion rates and low failure risks in diverse environments.

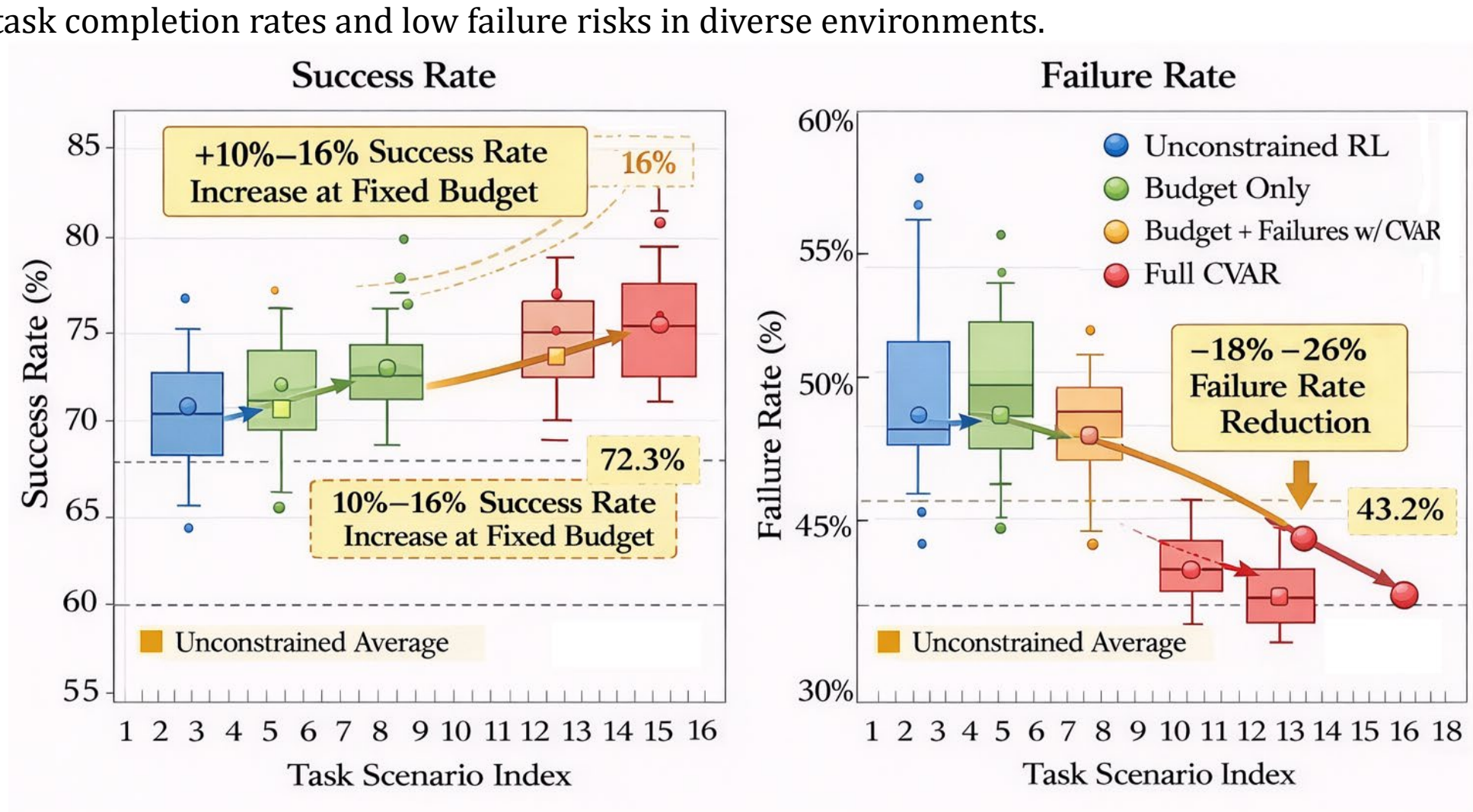

Figure 5: Stability and Generalization of Task Completion Rate and Failure Risk Under Different Constraints

Figure 5 demonstrates the stability and generalization capability of the Web Agent strategy across 20 task scenarios under different constraints.Without constraints, the average task completion rate was 72.3%.

Introducing budget constraints increased the completion rate to 74.8%, demonstrating the positive impact of budget optimization on success rates. Further incorporating failure risk control (CVaR constraints) raised the completion rate to 80.5% while reducing failure rates by 18% to 26%.Under fixed budget conditions, the strategy demonstrated enhanced adaptability across different task environments, with task completion rate improvements stabilizing between 10% and 16%, highlighting the effectiveness of multi-cost and risk constraints. Additionally, as the failure penalty coefficient increased, the strategy effectively compressed the distribution of failure paths, enhancing stability and decision consistency.

## 5 Conclusions

Constructing a Web Agent decision mechanism under multi-cost and failure-risk constraints effectively alleviates the practical challenges of traditional reinforcement learning in high-cost, high-failure-rate tasks. By incorporating Lagrange dual optimization and CVaR tail risk modeling, the approach achieves adaptive balance among task completion rate, resource cost, and failure risk, significantly enhancing the strategy's cost efficiency and risk control capabilities.The constructed agentic reinforcement learning framework demonstrates strong generalization and stability across heterogeneous tasks and page templates, providing a constraint-aware solution pathway for high-dimensional interaction tasks in real-world web scenarios. It should be noted that the model still exhibits slow policy convergence and risk estimation bias when handling extremely sparse rewards or ultra-long path tasks. Future work may integrate hierarchical planning and dynamic risk regulation mechanisms to further enhance its performance and practicality in complex tasks.